\documentclass[letterpaper]{article} % DO NOT CHANGE THIS
\usepackage{aaai2026}  % DO NOT CHANGE THIS
\usepackage{times}  % DO NOT CHANGE THIS
\usepackage{helvet}  % DO NOT CHANGE THIS
\usepackage{courier}  % DO NOT CHANGE THIS
\usepackage[hyphens]{url}  % DO NOT CHANGE THIS
\usepackage{graphicx} % DO NOT CHANGE THIS
\usepackage{natbib}  % DO NOT CHANGE THIS AND DO NOT ADD ANY OPTIONS TO IT
\usepackage{caption} % DO NOT CHANGE THIS AND DO NOT ADD ANY OPTIONS TO IT
\usepackage{algorithm}
\usepackage{algorithmic}

\usepackage{newfloat}
\usepackage{listings}

\usepackage{amsmath,amssymb,amsfonts}
\usepackage{booktabs}
\usepackage{multirow}

\DeclareCaptionStyle{ruled}{labelfont=normalfont,labelsep=colon,strut=off} % DO NOT CHANGE THIS
\floatstyle{ruled}
\newfloat{listing}{tb}{lst}{}
\floatname{listing}{Listing}
\title{Generating Attribute-Aware Human Motions from Textual Prompt}
\author{
    Xinghan Wang\textsuperscript{\rm 1}, Kun Xu, Fei Li\textsuperscript{\rm 2}, Cao Sheng\textsuperscript{\rm 2}, Jiazhong Yu\textsuperscript{\rm 2},Yadong Mu\textsuperscript{\rm 1}\thanks{Corresponding author.}
}
\affiliations{
    \textsuperscript{\rm 1}Peking University, China \\
    \textsuperscript{\rm 2}China Tower Corporation Limited\\
     \url{xinghan_wang@pku.edu.cn}, \url{myd@pku.edu.cn}
}
\usepackage{bibentry}
\begin{document}

\maketitle

\begin{abstract}
Text-driven human motion generation has recently attracted considerable attention, allowing models to generate human motions based on textual descriptions. However, current methods neglect the influence of human attributes—such as age, gender, weight, and height—which are key factors shaping human motion patterns. This work represents a pilot exploration for bridging this gap. We conceptualize each motion as comprising both attribute information and action semantics, where textual descriptions align exclusively with action semantics. To achieve this, a new framework inspired by Structural Causal Models is proposed to decouple action semantics from human attributes, enabling text-to-semantics prediction and attribute-controlled generation. The resulting model is capable of generating  attribute-aware motion aligned with the user's text and attribute inputs. For evaluation, we introduce a comprehensive dataset containing attribute annotations for text-motion pairs, setting the first benchmark for attribute-aware motion generation. Extensive experiments validate our model's effectiveness.
\end{abstract}

\section{Introduction}
\label{sec:intro}

\begin{figure}
    \centering
    \includegraphics[width=1.0\linewidth]{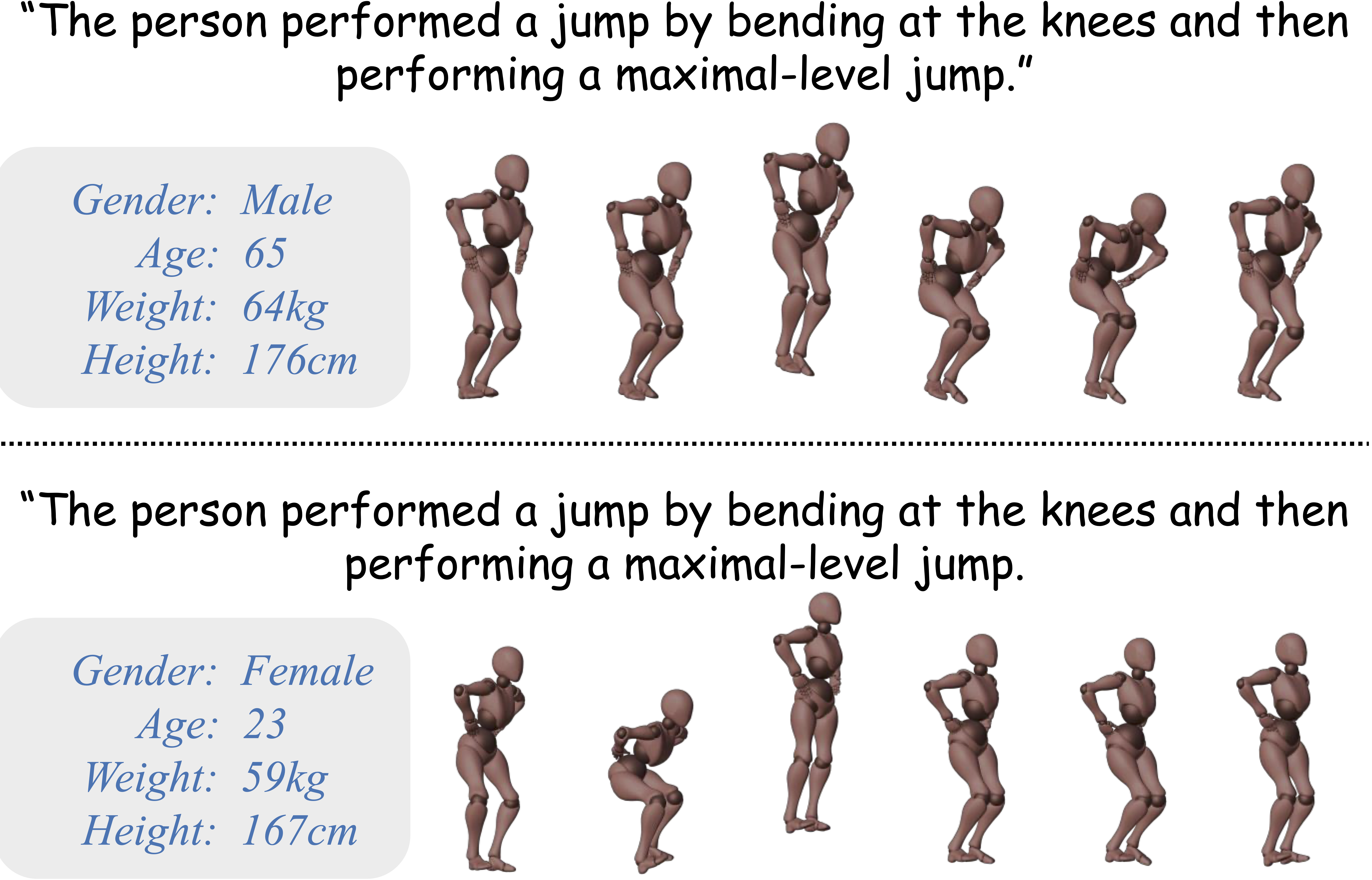}
    \caption{Data samples from HumanAttr dataset with text prompts and human attributes. Notice that the motion patterns of subjects with different attributes vary significantly.}
    \label{fig:motivation}
\end{figure}

In recent years, text-driven human motion generation has gained significant attention~\cite{humanml3d, zhang2022motiondiffuse, mld} in both research community and industry owing to its potential applications in video games, filmmaking, virtual reality, and robotics. The goal is to generate realistic human motion sequences based on given textual descriptions. However, human motion patterns are significantly influenced by human attributes such as age, gender, weight, and height. As illustrated in Figure~\ref{fig:motivation}, individuals with different attributes demonstrate distinct motion patterns. For instance, the movements of an elderly person and a teenager differ substantially. Therefore, incorporating human attributes into motion generation is crucial for producing realistic and context-aware movements. Also, users may need to generate motions that correspond to specific human attributes, demanding text-to-motion generation controlled by attributes. Despite this, none of the existing methods have addressed this issue.

Incorporating human attributes into text-driven motion generation faces two major challenges. The first is that a motion sequence consists of both \textit{action semantics} (\textit{e.g.}, walking, running) and \textit{human attributes} (\textit{e.g.}, gender, age), whereas textual descriptions usually focus solely on the action semantics aspect. Existing text-to-motion frameworks typically align text and motion in a shared space without distinguishing between semantic content and human attribute information, which may hinder the alignment. %For example, MotionCLIP~\cite{tevet2022motionclip} aligns textual and motion features in the CLIP latent space, while MotionGPT~\cite{motiongpt} obtains a unified token vocabulary for text and motion. 
This necessitates a new framework that can separate action semantics from human attributes, ensuring that texts are accurately aligned with semantic content. 

% Modified: pruned
The second challenge lies in the absence of a large-scale text-to-motion dataset that includes annotations for a broad spectrum of human attributes. Existing datasets either lack human attribute annotations entirely~\cite{humanml3d, lin2023motion, punnakkal2021babel, liang2023intergen, chen2024motionllm}, encompass subjects with only a small range of distinct attributes~\cite{plappert2016kit, jang2020etri, troje2002decomposing, ma2024nymeria,kim2021action}, or have very limited scale~\cite{sfu,  leightley2015benchmarking}. For example, the KIT dataset~\cite{plappert2016kit} includes attribute annotations for 4k motion sequences, but 90\% of the 55 subjects are aged between 18 and 45. 
%Similarly, the ETRI-LivingLab dataset~\cite{kim2021action} features 50 subjects, all aged between 64 and 80. 
%The lack of a proper benchmark has hindered the development of a robust attribute-aware motion generation model capable of generalizing to diverse textual descriptions and human attributes.
This data gap impedes the development of attribute-aware motion generation models capable of generalizing to diverse textual descriptions and human attributes.

% Our key insight is that human motion can be decomposed into two factors: action semantics and human attributes, where texts primarily focus on the semantics. 

This work represents a pilot exploration that incorporates human attributes to generate high-quality 3D human motion aligned with prompted text and attribute controls. To address the above-mentioned semantics-attribute coupling in human motions, we introduce \textbf{AttrMoGen} (\textbf{Attr}ibute-controlled \textbf{Mo}tion \textbf{Gen}erator). It comprises a Semantic-Attribute Decoupling VQVAE (Decoup-VQVAE) that extracts attribute-free semantic tokens from human motions, and a Semantics Generative Transformer that predicts these semantic tokens from textual prompts. To eliminate attribute information from the semantic tokens, we leverage Structural Causal Model (SCM)~\cite{pearl2018book}, formulating the problem as causal factor disentanglement. In particular, action semantics $S$ are causal factors determining target tokens $Y$, while human attributes $A$ are non-causal for $Y$ but crucial for constructing raw motion $X$. The causal model is trained using a causal information bottleneck, which disentangles $S$ from $A$ by limiting their mutual information $I(S;A)$. Additionally, a bottleneck term $-\lambda I(X; S)$ is used to restrict information flow from $X$ to $S$, ensuring that $S$ captures only the essential semantics. During inference, the Semantics Generative Transformer predicts semantic tokens from text input, which are combined with attributes input to generate the motion via the Decoup-VQVAE decoder.

We compiled a comprehensive dataset called \textbf{HumanAttr}, which includes texts and attribute annotations for each motion sequence. HumanAttr integrates data from multiple sources, resulting in a collection of 18.2k motion sequences from 640 subjects with a wide range of attributes. Example data entries are presented in Figure~\ref{fig:motivation}. Each attribute label consists of the subject's age and gender. A portion of the data ($\sim$ 74\%) also contains weight and height information, but lacks age diversity thus are not utilized in the main experiments. An additional experiment on this subset that contains weight and height is included in supplementary materials for reference. Extensive experiments are conducted on the dataset, evaluating a range of current representative text-to-motion models, including our proposed AttrMoGen model. The results demonstrate its ability to generate realistic motion that aligns with texts and attributes. 

% In summary, our contributions are threefold:
% \begin{itemize}
%     \item This work is the first attempt to incorporate human attributes into text-to-motion generation. We present a comprehensive text-to-motion dataset with attribute annotations whose subjects have diverse attributes.
%     \item A causality-based model that can decouple action semantics from human attributes is proposed, which enables text-to-semantics prediction and attribute-controlled generation.
%     \item Extensive experiments on HumanAttr validate the effectiveness of the proposed framework in generating attribute-aware human motion.
% \end{itemize}

% dataset

% experiments and conclusion.

% evaluated a number of models on HumanAttr dataset

% 解释一下：weight和height信息也有一部分，但是不全，数据集全部提供，但实验只用age和gender

\begin{table}[t]
    \centering
    %\small  
    \scalebox{0.6}{
    \begin{tabular}{lcccc}
        \toprule
        \textbf{Dataset} & \textbf{Subjects} & \textbf{Motions} & \textbf{Minutes} & \textbf{Age Range} \\
        \midrule
        BMLmovi~\cite{troje2002decomposing} & 86 & 1,801 & 161.8 & [17, 33] \\
        EKUT~\cite{plappert2016kit} & 4 & 348 & 30.1 & [24, 44] \\
        ETRI-Activity3D~\cite{jang2020etri} & 100 & 3,727 & 691.8 & [21, 88] \\
        ETRI-LivingLab~\cite{kim2021action} & 50 & 600 & 160.5 & [64, 80] \\
        K3Da~\cite{leightley2015benchmarking} & 54 & 236 & 34.5 & [18, 81] \\
        Kinder-Gator~\cite{aloba2018kinder} & 20 & 362 & 34.9 & [5, 32] \\
        KIT~\cite{plappert2016kit} & 55 & 4,231 & 463.1 & [15, 55] \\
        Nymeria~\cite{ma2024nymeria} & 264 & 6,850 & 552.4 & [18, 50] \\
        SFU~\cite{sfu} & 7 & 44 & 6.4 & [18, 30] \\
        \midrule
        \textbf{Total} & \textbf{640} & \textbf{18,199} & \textbf{2,135.4} & \textbf{[5, 88]} \\
        \bottomrule
    \end{tabular}}
    
    \caption{Statistics of the collected sub-datasets. %Note that for ~\cite{troje2002decomposing, plappert2016kit, sfu},  only entries with HumanML3D~\cite{humanml3d} text annotations are counted. 
    The statistics are gathered after data filtering.}
    \label{tab:HumanAttr}
\end{table}

\section{Related Work}
\label{sec:related}

% where 3D motion sequences are generated from provided text descriptions
% Modified: prune

\textbf{Human Motion Generation}. Generating realistic human motion that aligns with specific control signals has been a longstanding problem. A prominent focus is text-to-motion generation. Early works~\cite{tevet2022motionclip, petrovich2022temos, tmr, yu2024exploring} primarily focused on establishing a joint embedding space for motion sequences and texts. %For example, MotionCLIP~\cite{tevet2022motionclip} aligns human motion embeddings with the CLIP latent space using cosine similarity loss. %Similarly, TEMOS~\cite{petrovich2022temos} aligns motion and text by minimizing the distribution KL-divergence in some latent space.
Recent development has been dominated by autoregressive models and diffusion models. The former~\cite{humanml3d, huang2024controllable, humantomato} represent motions with discrete tokens and implement text-to-motion generation via next token prediction, %For instance, T2M-GPT~\cite{zhang2023t2m} employs a VQ-VAE to encode motion into discrete tokens, which can be generated autoregressively conditioned on text tokens. %MotionGPT~\cite{motiongpt} treats human motion as a foreign language, creating a joint token vocabulary of both motion and text. 
while diffusion models~\cite{jin2023act, yuan2023physdiff, zhang2023finemogen,wang2023fg, karunratanakul2023guided, mld, xie2023towards,xie2023omnicontrol, liang2024omg,huang2024stablemofusion} perform denoising in the motion space or latent space to generate natural and smooth human motions.
% ReMoDiffuse~\cite{zhang2023remodiffuse} further enhances the model by leveraging relevant knowledge from an auxiliary database to refine the denoising process. 
% Besides texts, motion can also be generated conditioned on other multimodal inputs, including action labels~\cite{action2motion, petrovich2021action, inr, actformer, multiact, case}, audio~\cite{fact, danceformer, bailando, edge, finedance, lodge, zhou2023ude, gong2023tm2d, zhang2024bidirectional, hoang2024motionmix, ling2024mcm, you2025momu}, history or future poses~\cite{aliakbarian2020stochastic, komura2017recurrent, harvey2020robust, kaufmann2020convolutional}, human trajectories~\cite{diffprior, gmd, xie2023omnicontrol, intercontrol, wan2023tlcontrol, pfnn, aamdm}, 3D-scenes or objects~\cite{cen2024generating, wang2024move, cha2024text2hoi, xu2023interdiff, kulkarni2024nifty, diller2024cg, yi2025generating, chen2024sitcom}, etc. % However, none has explored attribute-aware motion generation.
% Modified: add style motion related works
Recently, some studies~\cite{zhong2024smoodi, guo2024generative, kim2024most, song2024arbitrary} explore style-based motion models, emphasizing how an action is performed (\textit{e.g.}, proud, depressed, angry). Yet, style refers to subjective intent behind an action, while attributes determine the objective inherent biomechanics (\emph{e.g.}, child and the elderly naturally differ in stride length, joint range, and expressiveness). 
%Recently, some studies~\cite{xia2015realtime, zhong2024smoodi, guo2024generative, gupta2024d, zhang2024generative, kim2024most, song2024arbitrary} explore style-based motion models, emphasizing how an action is performed (\textit{e.g.}, proud, depressed, angry). In contrast, attributes denote people’s inherent and permanent characteristics (\textit{e.g.}, gender and age) which directly affect the biomechanics of their motion. %Although style labels may include terms like `childlike' or `old', they often indicate the performer is imitating the manner of movement of the child/elder. For instance, motions in widely used style-motion datasets~\cite{aberman2020unpaired, mason2022real} are performed by a single actor in various styles, which indicates attributes and styles are fundamentally different.
%Although some style labels like "childlike" may appear attribute-related, they typically depict performers are imitating specific motion patterns. The distinction is evident in widely-used style-motion datasets~\cite{aberman2020unpaired, mason2022real} where one single actor, characterized by certain attribute profile, performs actions in various styles.

% Recently, several works~\cite{guo2024momask, anisetty2024dynamic, li2024lamp, pinyoanuntapong2024controlmm} employ masking techniques similar to BERT~\cite{devlin2018bert}, where the model predicts randomly masked motion tokens conditioned on text input. 

% Modified: prune

\textbf{Text-motion Datasets}. %With the proliferation of large-scale mocap-based databases~\cite{ionescu2013human3, CMU_MoCap, shahroudy2016ntu, liu2019ntu, mandery2015kit, mahmood2019amass}, numerous text-motion datasets have been proposed. The conventional ones provide human motion data alongside corresponding sequence-level textual descriptions, primarily developed for text-to-motion and motion-to-text tasks. This includes the widely used benchmarks for human motion generation KIT~\cite{plappert2016kit} and HumanML3D~\cite{guo2022generating}. Subsequent works have advanced the field by incorporating multi-person scenarios~\cite{liang2023intergen, fang2024capturing, wang2024quo}, addressing long motion sequences~\cite{li2023sequential, han2023amd}, enhancing the quality of textual descriptions~\cite{tang2023flag3d, chen2024motionllm}, adding full-body text descriptions~\cite{lin2023motion}, and providing frame-level or segment-level annotations~\cite{punnakkal2021babel, wang2024text}.
With the proliferation of mocap-based databases~\cite{CMU_MoCap, shahroudy2016ntu, liu2019ntu, mandery2015kit, mahmood2019amass}, numerous text-motion datasets have been proposed. Conventional ones like KIT and HumanML3D provide motion with corresponding sequence-level textual descriptions, primarily developed for text-to-motion task. Subsequent works have advanced the field by incorporating multi-person scenarios~\cite{liang2023intergen, fang2024capturing}, addressing long motion sequences~\cite{li2023sequential, han2023amd}, enhancing the quality of text descriptions~\cite{tang2023flag3d, chen2024motionllm}, adding full-body text descriptions~\cite{lin2023motion}, and providing frame-level or segment-level annotations~\cite{punnakkal2021babel, wang2024text}.
% In addition to datasets with solely textual annotations, recent studies have introduced diverse input modalities beyond text. For instance, some works~\cite{wang2022humanise, zhao2024m, jiang2023full} combine object or 3D scene information with text-motion datasets, enabling generation in the context of human-object or human-scene interactions. Additionally, some studies integrate audio inputs, such as music~\cite{zhang2024large, li2021ai} and speech~\cite{zhang2024large, yoon2020speech, liu2022beat}, to control the generation process. Several motion datasets~\cite{sfu, plappert2016kit, troje2002decomposing, ma2024nymeria, jang2020etri, kim2021action} provide the attribute information of each subject. However, they often have limited data scale or attribute range. For instance, 90\% of the subjects in KIT~\cite{plappert2016kit} are aged between 18 and 45. This indicates that current datasets lack large-scale collections that encompass a wide range of human attributes.
%In addition to datasets with solely text annotations, recent studies introduced diverse annotations' modalities, such as object or 3D scene~\cite{wang2022humanise, zhao2024m, jiang2023full}, music~\cite{zhang2024large, li2021ai} and speech~\cite{zhang2024large, yoon2020speech, liu2022beat}. 
Several motion datasets~\cite{sfu, plappert2016kit, troje2002decomposing, ma2024nymeria, jang2020etri, kim2021action} provide the attribute information of each subject, yet are limited in data scale or attribute range. For instance, 90\% of the subjects in KIT are aged between 18 and 45. This indicates the lack of large-scale collections that encompass a wide range of human attributes.

\begin{figure}[t]
    \centering
    \includegraphics[width=\columnwidth]{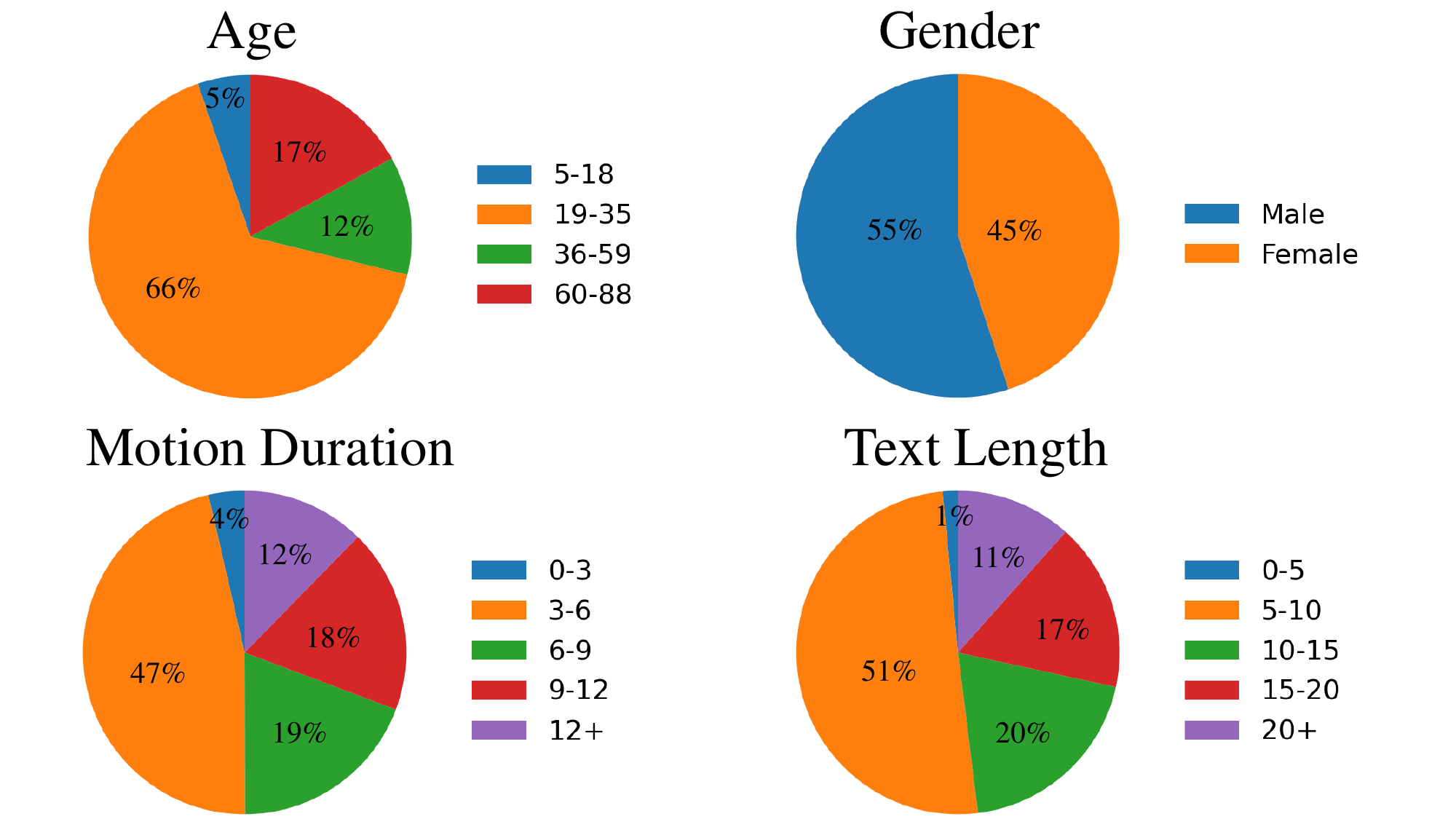}
    \caption{Statistics of age, gender, motion duration (in seconds), and text length (in words) of the HumanAttr dataset.}
    \label{fig:dataset}
\end{figure}

\section{The HumanAttr Dataset}
\label{sec:dataset}
\textbf{Dataset Collection}. Existing benchmark datasets for human motion generation such as HumanML3D and KIT primarily consist of motion sequences and corresponding text descriptions, lacking the annotations for human attributes. To address this gap, we compile a comprehensive dataset that includes data entries of (motion, text, attribute). The attribute labels contain age and gender for each data entry, and a portion of the data ($\sim$74\%) also have weight and height information. The dataset, named HumanAttr, integrates data from various sources, annotated with diverse human attributes and textual descriptions. As illustrated in Table~\ref{tab:HumanAttr}, HumanAttr consists of nine sub-datasets that encompass a variety of subjects with varying attributes. In total, it contains 18.2k motion sequences from 640 subjects, with ages ranging from 5 to 88 years and balanced gender. Figure~\ref{fig:dataset} presents the distribution of age, gender, motion duration, and text length within the dataset.

\textbf{Dataset Processing}. This section provides a brief introduction to the pipeline for processing HumanAttr. Initially, data from different sources are converted into a 3D coordinates representation of $L \times V \times 3$, where $L$ denotes the sequence length and $V$ is the number of markers. Since the number of markers varies across different data sources, we adopt the method from \cite{li2024isolated} and use gradient descent optimization to fit the marker data into unified parameterized SMPL model~\cite{loper2023smpl}. To make the dataset compatible with previous work in the field, processing protocol from HumanML3D~\cite{humanml3d} is then applied, resulting in a 263-dimensional input vector.

However, some of the motion data exhibited serious jittering issues caused by the Kinect system. To address this, additional processing is applied to specific sub-datasets. Motions with excessive jitter or outlier coordinates are discarded, and a 1-D Gaussian denoising is applied to enhance motion quality. The dataset is further augmented by mirroring the motion data following the approach in HumanML3D to increase the training samples and improve model generalization. For text annotations, all sub-datasets include either textual descriptions or action category labels. Action labels are expanded into full text descriptions using the information from the original database, ensuring each motion entry has a meaningful text.

\section{Our Proposed Method}
\label{sec:method}
Our goal is to generate 3D human motion that aligns with both a given textual description and human attributes control. The core observation is that human motion patterns can be decomposed into two factors: action semantics and human attributes. Since textual descriptions primarily focus on the semantics, existing methods that directly align motion with text annotations may hinder the alignment. Moreover, current methods lack the capability to generate attribute-aware motion based on attribute control input.

\begin{figure*}[t]
    \centering
    \includegraphics[width=0.95\linewidth]{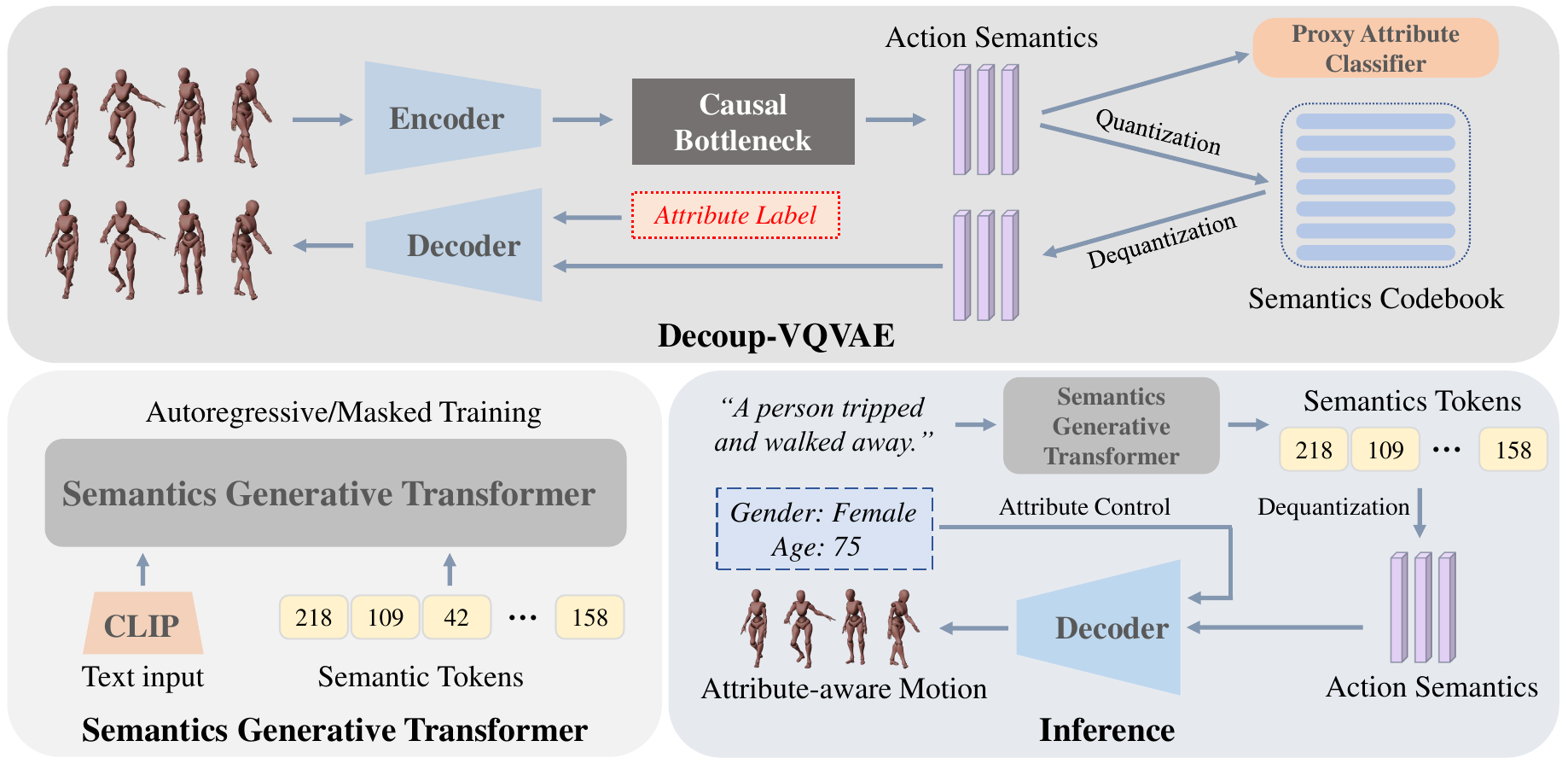}
    \caption{Overall architecture of our proposed AttrMoGen. The encoder of Decoup-VQVAE uses a causal information bottleneck to decouple action semantics from human attributes, producing attribute-free semantic tokens. The decoder then reconstructs motion from these semantic tokens and attribute labels. The Semantics Generative Transformer predicts semantic tokens from textual input, which are subsequently combined with attribute inputs to generate attribute-aware human motions during inference.}
    \label{fig:overall}
\end{figure*}

To tackle this challenge, we propose an architecture composed of two main components. The overall architecture is illustrated in Figure~\ref{fig:overall}. The first component is a Semantic-attribute Decoupling VQVAE (Decoup-VQVAE), inspired by Structural Causal Model (SCM)~\cite{pearl2018book}. Specifically, the encoder removes attribute information from the raw motion with the help of a causal information bottleneck, resulting in attribute-free semantic tokens via vector quantization. The decoder can then reconstruct the original motion using both the semantic tokens and attribute labels. The second component is a Semantics Generative Transformer, which predicts semantic tokens from text by a generative transformer. During inference, semantic tokens are generated from textual input and combined with user-defined attributes to generate attribute-aware human motion. 

%Details of the framework will be elaborated in the following sections.

\begin{figure}[t]
    \centering
    \includegraphics[width=0.7\linewidth]{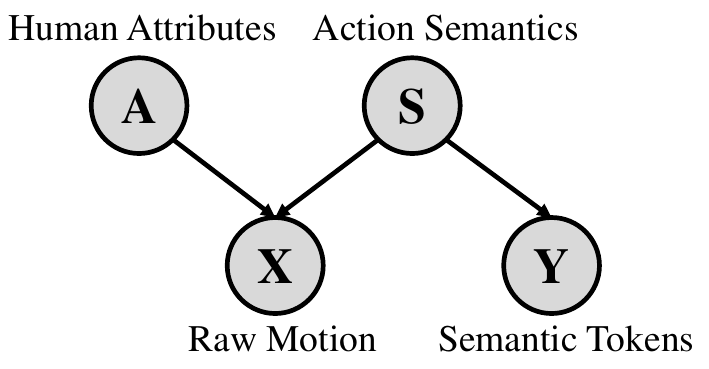}
    \caption{Structural Causal Model for our Decoup-VQVAE. Our objective is to learn an encoder capable of decoupling the $Y$-causative action semantics $S$ from raw motion $X$.}
    \label{fig:scm}
\end{figure}

\subsection{Semantic-attribute Decoupling VQVAE}
To obtain attribute-free semantic tokens, we employed Structural Causal Model (SCM) to formulate the problem as causal factor disentanglement. Specifically, let $X$ represent the raw motion and $Y$ represent the desired semantic tokens. The action semantics $S$ are the causal factors that determine $Y$. Human attributes, denoted as $A$, are non-causal factors that are independent of $Y$, but essential in constructing the raw motion $X$. This leads to our causal formulation, with the SCM illustrated in Figure~\ref{fig:scm}. Our aim is to learn an encoder $S = f(X)$ that can decouple the $Y$-causative action semantics $S$ from raw motion $X$, and a decoder $\hat{X} = g(S, A)$ that can reconstruct the motion $X$ from $S, A$. Ideally, $f$ should clearly separate $S$ from $A$, retaining only the essential semantic information of the raw motion. Inspired by~\cite{zhang2024causaldiff, hua2022causal}, we introduce a causal information bottleneck (CIB) to tackle this challenge. Formally, The CIB objective function is defined as
\begin{multline}
CIB(X,Y,S,A) = I(X;S,A) + I(Y;S) \\
- I(S;A) - \lambda I(X;S),
\end{multline}
where $I(;)$ denotes the mutual information and $\lambda$ is the bottleneck weight. \cite{zhang2024causaldiff} proves that maximizing the CIB objective is equivalent to maximizing the mutual information $I(X, Y; S, A)$ between the latent factors $S$ and $A$ and the variables $X$ and $Y$ with an information bottleneck. The role of each term and our customized implementation will be demonstrated below.

\paragraph{Decoupling.} The term $-I(S; A)$ restricts the mutual information between $S$ and $A$, which plays a crucial role in disentangling the causal factor $S$ from the non-causal factor $A$. In our case, note that the mutual information has the following upper bound:
\begin{align}
    I(S;A)  &= H(A) - H(A|S) \\
    &\le log|A| - \mathbb{E}_{s \sim p(S)} H(A|S=s),
\end{align}
where $H$ denotes the entropy and $|A|$ denotes the total size of the attribute space. Since the attribute set is finite, $\log |A|$ is a constant. Given a batch of samples $\{s_i\}_{i=1}^B$, the above upper bound of mutual information can be estimated as:
\begin{align}
    \hat{I}(S;A)& = const - \sum_{i=1}^B H(A|S=s_i) \\
    &= const + \sum_{i=1}^B \sum_{a \in \mathcal{A}} p(a|s_i) \log p(a|s_i),
    \label{eq:entropy}
\end{align}
where $\mathcal{A}$ is the attribute label set. The problem reduces to modeling the conditional distribution $p(a|s_i)$. To this end, a proxy attribute classification network $h$ is introduced, whose goal is to classify the human attribute $A$ from the semantic embedding $S$. The output score of $h$ then indicates the conditional probability: $p(A|s_i) = h(s_i)$, which enables the computation of the following loss function through Equation~\ref{eq:entropy}:
\begin{equation}
    \mathcal{L}_{entropy} = -\sum_{i=1}^B H(A|S=s_i).
\end{equation}
Minimizing $\mathcal{L}_{entropy}$ reduces the mutual information $I(S;A)$ between $S$ and $A$, eliminating attribute information from $S$ and achieving effective decoupling. During training, the proxy attribute classifier $h$ is updated alternately with the encoder $f$ and decoder $g$, supervised by the cross-entropy loss $\mathcal{L}_{CE}$ with ground truth attribute labels.

\paragraph{Bottleneck.} The information bottleneck term $-\lambda I(X; S)$ limits the information flow from $X$ to $S$, ensuring that $S$ contains only essential semantics without excessive irrelevant information, fostering a robust and compact representation~\cite{tishby2000information}. The mutual information $I(X; S)$ can be measured using the Kullback-Leibler (KL) divergence:
% \begin{align}
%     I(X;S) &= \mathbb{E}_{x \sim p(X)} \mathcal{D}_{KL}(p(S|x) \| p(S)),
% \end{align}
\begin{align}
    I(X;S) &= \mathbb{E}_{x \sim p(X)} \mathcal{D}_{KL}(p(S|x) \| p(S)) \\
            &= \mathbb{E}_{x \sim p(X)} \mathcal{D}_{KL}(p(S|x) \| \mathbb{E}_{x' \sim p(X)} p(S|x')) \\
            & \le \mathbb{E}_{x \sim p(X)} \mathbb{E}_{x' \sim p(X)} \mathcal{D}_{KL}(p(S|x) \|  p(S|x')).
\end{align}
The inequality holds according to Jensen's inequality as the KL divergence is convex \textit{w.r.t.} the second argument, and a rigorous proof is provided in the supplementary. The objective is basically alleviating the impact of $X$ on the posterior distribution of $S$. To achieve this, we minimize the following objective:
\begin{equation}
    D(X;X^-) = \mathcal{D}_{KL}(p(S|X) \| p(S|X^-)),
    \label{eq:kl}
\end{equation}
where $X^-$ is the counterfactual motion that shares the same semantics with the original $X$ but has different attributes. Note that our decoder $g$ can generate motion based on specific semantics and any given attribute control, therefore $X^{-}$ can be derived through
\begin{equation}
    X^- = g(S, A^-),
\end{equation}
where $A^{-}$ is the counterfactual attribute obtained by randomizing the original attribute $A$. Suppose $\boldsymbol{X}$ is a batch of samples and $\boldsymbol{X}^-$ is their corresponding counterfactuals. As the encoder $S = f(X)$ serves as an estimator for the distribution $p(S|X)$, the KL-divergence in Equation~\ref{eq:kl} can be minimized by aligning $f(\boldsymbol{X})$ and $f(\boldsymbol{X}^-)$. To achieve this, we define the following similarity matrix analogous to~\cite{lv2022causality}:
% \begin{equation}
%     \boldsymbol{C}(\boldsymbol{S}, \boldsymbol{S}^-)_{ij} = \frac{< \boldsymbol{S}[:, i], \boldsymbol{S}^-[:, j] >}{\|\boldsymbol{S}[:, i]\|_F \|\boldsymbol{S}[:, j]\|_F}, 
% \end{equation}
\begin{equation}
    \boldsymbol{\tilde{D}}(\boldsymbol{X}, \boldsymbol{X}^-)_{ij} = Cosine(\boldsymbol{S}[:, i], \boldsymbol{S}^-[:, j]),
\end{equation}
where $\boldsymbol{S} = f(\boldsymbol{X}) \in \mathbb{R}^{B \times D}$, $\boldsymbol{S}^- = f(\boldsymbol{X}^-) \in \mathbb{R}^{B \times D}$, %$< , >$ denotes the inner product.
$\boldsymbol{\tilde{D}} \in \mathbb{R}^{D \times D}$, $Cosine$ denotes cosine similarity. This brings the following loss function, which enforces $\boldsymbol{S}$ and $\boldsymbol{S}^-$ to be closer while preserving independence between channels, fostering feature space diversity:
\begin{equation}
    \mathcal{L}_{bottleneck} = \|\boldsymbol{\tilde{D}}(\boldsymbol{X}, \boldsymbol{X}^-) - \boldsymbol{I}\|_F^2,
\end{equation}
where $\boldsymbol{I}$ denotes the identity matrix. A detailed illustration of this workflow is in Algorithm~\ref{alg:algo}.

\begin{algorithm}[t]
\caption{Optimization of Decoup-VQVAE}
\label{alg:algo}

\begin{algorithmic}
\STATE \textbf{for} each training iteration \textbf{do}
\STATE \quad 1. Sample a batch $\{(X_i, A_i)\}_{i=1}^{B}$ from dataset.
\STATE \quad 2. Calculate semantic embedding $S_i=f(X_i)$.
\STATE \quad 3. Randomize the attribute label $A_i$ into $A_i^-$.
\STATE \quad 4. Obtain counterfactual motion $X_i^- = g(S_i, A_i^-)$.
\STATE \quad 5. Update encoder $f(S|X)$ and decoder $g(X|S, A)$ by minimizing $\mathcal{L}_{vqvae} + \alpha \mathcal{L}_{entropy} + \lambda \mathcal{L}_{bottleneck}$.
\STATE \quad 6. Update proxy attribute classifier $h(A|S)$ by minimizing $\mathcal{L}_{CE}$.
\STATE \textbf{end for}
\end{algorithmic}
\end{algorithm}

\paragraph{Reconstruction.} 
The remaining terms, $I(X; S, A) + I(Y; S)$, ensure that the information contained in $S$ and $A$ are sufficient to reconstruct $X$, while also guarantee that $S$ is informative enough to deduce $Y$. Note the following lower bound derived by~\cite{zhang2024causaldiff}:
\begin{equation}
    I(X; S, A) \ge \mathbb{E}_{p(X, S, A)} \log p_\theta(X | S, A),
\end{equation}
where $p_\theta$ denotes the decoder $g$. This implies that $I(X; S, A)$ can be addressed using a reconstruction loss
\begin{equation}
    \mathcal{L}_{rec} = |X - \hat{X}|_F^2 = |X - g(S, A)|_F^2.
\end{equation}
The second term $I(Y; S)$ ensures the deduction from the causal factor $S$ to the target semantic token $Y$, which is, by definition, the quantization process from the semantic embedding to the codebook index.
Note that $I(Y; S) = H(Y) - H(Y|S)$, where $H(Y)$ reflects the full exploitation of the codebook and $-H(Y|S)$ demands precise indexing from $S$ to $Y$. This can be addressed by the loss $\mathcal{L}_{commit} + \mathcal{L}_{embed}$ from VQVAE~\cite{van2017neural}. In summary, the overall loss function is formulated as 
\begin{equation}
    \mathcal{L}_{overall} = \mathcal{L}_{vqvae} + \alpha \mathcal{L}_{entropy} + \lambda \mathcal{L}_{bottleneck},
    \label{eq:loss}
\end{equation}
where $\mathcal{L}_{vqvae} = \mathcal{L}_{rec} + \mathcal{L}_{commit} + \mathcal{L}_{embed}$ is the conventional VQVAE loss and $\alpha, \lambda$ are weight coefficients. 

\subsection{Semantics Generative Transformer}
To predict semantic tokens from text, a Semantics Generative Transformer is employed, whose architecture can be either a GPT-like autoregressive model~\cite{zhang2023t2m, humantomato} or a masked transformer~\cite{guo2024momask}. We adopt MoMask~\cite{guo2024momask} due to its strong performance and efficiency. During training, the model aims to predict randomly masked semantic tokens conditioned on the CLIP~\cite{radford2021learning} features of the textual input, following a BERT-like scheme~\cite{devlin2018bert}. At inference time, semantic tokens are first predicted from the text using the Semantics Generative Transformer. These tokens, combined with attribute inputs (\emph{e.g.}, age, gender), are then used to generate the motion via the Decoup-VQVAE decoder. Figure~\ref{fig:overall} presents a schematic illustration.

\begin{table}[t]

  \centering
  \Large
  \scalebox{0.56}
{
  \begin{tabular}{lccccc}
    \toprule[1.25pt]
\multirow{2}{*}{{Methods}} &\multicolumn{3}{c}{{R-Precision $\uparrow$}} & \multirow{2}{*}{{FID $\downarrow$}} & \multirow{2}{*}{{MM-Dist $\downarrow$}} \\
\cmidrule(rl){2-4}
  & Top-1 & Top-2 & Top-3 \\ 
    \midrule
    MoMask &0.685$^{\pm.0.003}$ & 0.864$^{\pm.0.002}$ & 0.925$^{\pm.0.001}$ & 0.245$^{\pm.0.009}$ & 2.602$^{\pm.0.009}$ \\
     \quad w/ attr test  & 0.603$^{\pm.0.002}$ & 0.781$^{\pm.0.002}$ & 0.853$^{\pm.0.002}$ & 0.957$^{\pm.0.019}$ & 3.815$^{\pm.0.014}$ \\
     \quad w/ attr train  &0.689$^{\pm.0.002}$ & 0.872$^{\pm.0.002}$ & 0.933$^{\pm.0.002}$& 0.203$^{\pm.0.007}$ &2.518$^{\pm.0.012}$ \\
    \midrule
    \midrule
    w/o entropy  &0.686$^{\pm.0.003}$ & 0.867$^{\pm.0.002}$ &0.930$^{\pm.0.001}$ & 0.489$^{\pm.0.008}$ & 2.523$^{\pm.0.014}$ \\
    w/o bottleneck  &  0.686$^{\pm.0.003}$ &   0.870$^{\pm.0.002}$& 0.933$^{\pm.0.001}$  &   0.184$^{\pm.0.006}$ &  2.486$^{\pm.0.011}$ \\
    \midrule
    
    $\lambda$ = 0.25  & 0.698$^{\pm.0.002}$ & 0.881$^{\pm.0.002}$ & \textbf{0.942}$^{\pm.0.002}$& 0.098$^{\pm.0.004}$ &  2.326$^{\pm.0.012}$ \\
         
    $\lambda$ = 1  & 0.701$^{\pm.0.002}$ & 0.881$^{\pm.0.002}$ & 0.941$^{\pm.0.001}$ &  \textbf{0.088}$^{\pm.0.003}$ & 2.332$^{\pm.0.007}$ \\
    
    \midrule

    $\alpha$ = 0.005  & 0.691$^{\pm.0.002}$ & 0.869$^{\pm.0.001}$ & 0.931$^{\pm.0.001}$& 0.236$^{\pm.0.007}$ &  2.481$^{\pm.0.016}$ \\
    
    $\alpha$ = 0.02  & 0.697$^{\pm.0.002}$ & 0.878$^{\pm.0.002}$ & 0.939$^{\pm.0.001}$& 0.139$^{\pm.0.004}$ &  2.364$^{\pm.0.012}$ \\

    \midrule

    \textbf{AttrMoGen} &  \textbf{0.705}$^{\pm.0.002}$ &   \textbf{0.882}$^{\pm.0.002}$& 0.940$^{\pm.0.001}$  &  0.089$^{\pm.0.003}$&  \textbf{2.266}$^{\pm.0.012}$  \\
  \bottomrule
  
  \end{tabular}}
    \caption{
    Results of ablation studies.
    ``$\uparrow$'' denotes that higher is better.
    ``$\downarrow$'' denotes that lower is better. The default settings for AttrMogen are $\lambda = 0.5$ and $\alpha = 0.01$.}
    \label{tab:ablation}
\end{table}

\begin{table*}[t]
  \centering
  \scalebox{0.81}
{
  \begin{tabular}{lccccccc}
    \toprule[1.25pt]
\multirow{2}{*}{{Methods}} &\multicolumn{3}{c}{{R-Precision $\uparrow$}} & \multirow{2}{*}{{FID $\downarrow$}} & \multirow{2}{*}{{MM-Dist $\downarrow$}} & \multirow{2}{*}{{Diversity $\rightarrow$}} & \multirow{2}{*}{{MModality $\uparrow$}}\\
\cmidrule(rl){2-4}
  & Top-1 & Top-2 & Top-3 \\ 
    \midrule
    Real motion & 0.750$^{\pm.0.002}$ & 0.913$^{\pm.0.002}$ &0.964$^{\pm.0.001}$ &0.001$^{\pm.0.000}$  & 1.793$^{\pm.0.004}$ & 19.259$^{\pm.0.152}$ & - \\ 
    \midrule
    TM2T~\cite{guo2022tm2t} &  0.479$^{\pm.0.002}$ & 0.686$^{\pm.0.002}$ & 0.800$^{\pm.0.002}$  & 1.491$^{\pm.0.033}$ & 3.787$^{\pm.0.014}$ & 18.598$^{\pm.0.142}$ & 2.841$^{\pm.0.085}$ \\
    T2M~\cite{humanml3d} & 0.592$^{\pm.0.002}$ & 0.778$^{\pm.0.002}$ &0.859$^{\pm.0.001}$ & 1.909$^{\pm.0.031}$ &3.827$^{\pm.0.018}$  & 18.856$^{\pm.0.217}$ & 2.627$^{\pm.0.133}$\\
    MotionDiffuse~\cite{zhang2022motiondiffuse} & 0.670$^{\pm.0.002}$ & 0.857$^{\pm.0.002}$ &0.928$^{\pm.0.001}$ & 0.416$^{\pm.0.011}$ & 2.704$^{\pm.0.015}$ & 18.968$^{\pm.0.150}$ & 2.435$^{\pm.0.119}$\\
    MDM~\cite{mdm} & 0.330$^{\pm.0.006}$&  0.479$^{\pm.0.009}$ & 0.565$^{\pm.0.007}$&  1.581$^{\pm.0.361}$& 9.482$^{\pm.0.105}$ & 16.294$^{\pm.0.296}$ & \textbf{4.853}$^{\pm.0.084}$\\
    T2M-GPT~\cite{zhang2023t2m} & 0.661$^{\pm.0.002}$ & 0.824$^{\pm.0.002}$ & 0.889$^{\pm.0.002}$&  0.279$^{\pm.0.017}$& 3.203$^{\pm.0.022}$ & 19.184$^{\pm.0.190}$ & 3.170$^{\pm.0.162}$ \\
    MLD~\cite{mld} & 0.660$^{\pm.0.002}$ & 0.850$^{\pm.0.001}$ & 0.922$^{\pm.0.001}$ & 0.349$^{\pm.0.011}$ & 2.673$^{\pm.0.013}$ & 19.576$^{\pm.0.239}$  & 1.938$^{\pm.0.086}$ \\        
    GenMoStyle~\cite{guo2024generative} & 0.680$^{\pm.0.001}$ & 0.861$^{\pm.0.002}$ & 0.925$^{\pm.0.001}$ & 0.332$^{\pm.0.007}$ & 2.649$^{\pm.0.012}$ & 19.118$^{\pm.0.206}$  & 1.588$^{\pm.0.074}$ \\
    MoMask~\cite{guo2024momask} &0.685$^{\pm.0.003}$ & 0.864$^{\pm.0.002}$ & 0.925$^{\pm.0.001}$ & 0.245$^{\pm.0.009}$ & 2.602$^{\pm.0.009}$ &18.981$^{\pm.0.132}$  & 1.438$^{\pm.0.084}$ \\
    \midrule
     \textbf{AttrMoGen} & \textbf{0.705}$^{\pm.0.002}$ &  \textbf{0.882}$^{\pm.0.002}$&\textbf{0.940}$^{\pm.0.001}$  & \textbf{0.089}$^{\pm.0.003}$& \textbf{2.266}$^{\pm.0.012}$ & \textbf{19.268}$^{\pm.0.212}$ & 1.250$^{\pm.0.076}$\\
  \bottomrule
  
  \end{tabular}}
    \caption{
    Performance of representative existing text-to-motion methods as well as our proposed AttrMoGen which incorporates attribute information on the HumanAttr test set.
    ``$\uparrow$'' denotes that higher is better.
    ``$\downarrow$'' denotes that lower is better.
    ``$\rightarrow$'' denotes that results are better when closer to the real motion.}
    \label{tab:overall}
\end{table*}

\section{Experiments}
\label{sec:experiment}

\begin{table*}[t]
  \centering
  \scalebox{0.81}
  {
    \begin{tabular}{cccccccccc}
      \toprule[1.25pt]
      \multirow{2}{*}{{Attribute}} &\multirow{2}{*}{{Group}} & \multirow{2}{*}{{Method}} & \multicolumn{3}{c}{{R-Precision $\uparrow$}} & \multirow{2}{*}{{FID $\downarrow$}} & \multirow{2}{*}{{MM-Dist $\downarrow$}} &Diversity $\rightarrow$ \\
      \cmidrule(rl){4-6}
       & & & Top-1 & Top-2 & Top-3 & & &  Generated / Real\\
      \midrule
      \multirow{8}{*}{Age}  & \multirow{2}{*}{5-18} & MoMask & 0.511$^{\pm.0.010}$ &0.733$^{\pm.0.010}$ & 0.835$^{\pm.0.008}$ &\textbf{0.736}$^{\pm.0.094}$ &2.630$^{\pm.0.042}$ & 14.144$^{\pm.0.229}$ / 14.552$^{\pm.0.200}$ \\

       & & AttrMoGen & \textbf{0.544}$^{\pm.0.007}$ & \textbf{0.752}$^{\pm.0.009}$ & \textbf{0.842}$^{\pm.0.005}$ & 0.778$^{\pm.0.079}$ & \textbf{2.487}$^{\pm.0.038}$ &  \textbf{14.500}$^{\pm.0.179}$ / 14.552$^{\pm.0.200}$ \\
        \cmidrule(rl){2-9}
       
       & \multirow{2}{*}{19-35} & MoMask & 0.684$^{\pm.0.004}$ & 0.858$^{\pm.0.002}$& 0.918$^{\pm.0.002}$  & 0.240$^{\pm.0.011}$ &2.634$^{\pm.0.019}$  &   18.571$^{\pm.0.133}$ / 18.836$^{\pm.0.118}$ \\
       
       & & AttrMoGen & \textbf{0.702}$^{\pm.0.003}$ &  \textbf{0.876}$^{\pm.0.002}$& \textbf{0.932}$^{\pm.0.002}$  & \textbf{0.090}$^{\pm.0.003}$ & \textbf{2.349}$^{\pm.0.012}$ &  \textbf{18.708}$^{\pm.0.205}$ / 18.836$^{\pm.0.118}$ \\
       \cmidrule(rl){2-9}

       & \multirow{2}{*}{36-59} & MoMask & 0.445$^{\pm.0.007}$ &0.643$^{\pm.0.005}$ & 0.744$^{\pm.0.006}$ &1.944$^{\pm.0.092}$ & 3.762$^{\pm.0.037}$ &  14.939$^{\pm.0.129}$ / 14.508$^{\pm.0.156}$\\
       
       & & AttrMoGen & \textbf{0.483}$^{\pm.0.007}$ &\textbf{0.677}$^{\pm.0.008}$ & \textbf{0.781}$^{\pm.0.006}$ & \textbf{0.669}$^{\pm.0.039}$&\textbf{3.185}$^{\pm.0.035}$  & \textbf{14.873}$^{\pm.0.114}$ / 14.508$^{\pm.0.156}$ \\

        \cmidrule(rl){2-9}

      & \multirow{2}{*}{60-88} & MoMask & 0.375$^{\pm.0.004}$ & 0.597$^{\pm.0.003}$ & 0.731$^{\pm.0.003}$ & 0.145$^{\pm.0.014}$&1.702$^{\pm.0.022}$ &  12.175$^{\pm.0.131}$ / 11.953$^{\pm.0.154}$\\
       
       & & AttrMoGen & \textbf{0.393}$^{\pm.0.003}$ &\textbf{0.618}$^{\pm.0.003}$ & \textbf{0.755}$^{\pm.0.004}$ &\textbf{0.097}$^{\pm.0.009}$ & \textbf{1.209}$^{\pm.0.017}$ &  \textbf{11.915}$^{\pm.0.142}$ / 11.953$^{\pm.0.154}$\\
        \midrule \midrule
       
       \multirow{4}{*}{Gender}  & \multirow{2}{*}{male} & MoMask & 0.672$^{\pm.0.003}$ &0.851$^{\pm.0.002}$ &  0.914$^{\pm.0.001}$& 0.277$^{\pm.0.010}$& 2.710$^{\pm.0.016}$  & 18.664$^{\pm.0.194}$ / 19.092$^{\pm.0.172}$ \\
       
       & & AttrMoGen & \textbf{0.701}$^{\pm.0.003}$ &\textbf{0.877}$^{\pm.0.002}$ &  \textbf{0.934}$^{\pm.0.002}$& \textbf{0.097}$^{\pm.0.006}$&\textbf{2.340}$^{\pm.0.020}$  & \textbf{18.910}$^{\pm.0.179}$ / 19.092$^{\pm.0.172}$ \\

        \cmidrule(rl){2-9}
        
       & \multirow{2}{*}{female} & MoMask &  0.681$^{\pm.0.003}$ & 0.871$^{\pm.0.003}$&  0.934$^{\pm.0.002}$ &0.246$^{\pm.0.010}$ & 2.468$^{\pm.0.012}$ & 19.232$^{\pm.0.194}$ / 19.433$^{\pm.0.223}$ \\
       
       & &  AttrMoGen & \textbf{0.694}$^{\pm.0.004}$ & \textbf{0.879}$^{\pm.0.003}$& \textbf{0.941}$^{\pm.0.002}$ &\textbf{0.118}$^{\pm.0.006}$ & \textbf{2.176}$^{\pm.0.018}$  & \textbf{19.615}$^{\pm.0.148}$ / 19.433$^{\pm.0.223}$ \\
      \bottomrule

    \end{tabular}
  }
    \caption{
    Performance comparison on different attribute groups.
    ``$\uparrow$'' denotes that higher is better.
    ``$\downarrow$'' denotes that lower is better.
    ``$\rightarrow$'' denotes that results are better when closer to the real motion.}
    \label{tab:group}
\end{table*}

\paragraph{Implementation and Evaluation Details.} 
For the model architecture, MoMask~\cite{guo2024momask} is adopted due to its superior performance and efficiency. To ensure a fair comparison, all experimental settings and hyperparameters strictly follow those in MoMask. The weight coefficients $\alpha, \lambda$ of Decoup-VQVAE in Equation~\ref{eq:loss} are set to $0.01, 0.5$ empirically. For attribute control, we categorize the attributes into discrete labels to accommodate sub-datasets that only provide age groups (e.g., Nymeria) and enable in-depth analyses based on categorization. Gender is labeled 0 (male) or 1 (female); age is grouped as 0 (5–18), 1 (19–35), 2 (36–59), and 3 (60–88). One-hot encodings of these labels are fused via MLPs, then concatenated with semantic embeddings for decoding. For the evaluation metrics, we follow the standard protocol~\cite{humanml3d} and report the following metrics: FID, R-Precision, Multimodal Distance (MM-Dist), Diversity and Multimodality.

%Frechet Inception Distance (FID)~\cite{heusel2017gans}, R-Precision, Multimodal Distance (MM-Dist), Diversity and Multimodality. The feature extractor used for evaluation is pre-trained on the HumanAttr dataset, following the same method as in~\cite{humanml3d}. Detailed introduction of these metrics is included in the supplementary.

% Modified: add hyperparameters

\begin{figure*}[t]
    \centering
    \includegraphics[width=1.0\textwidth]{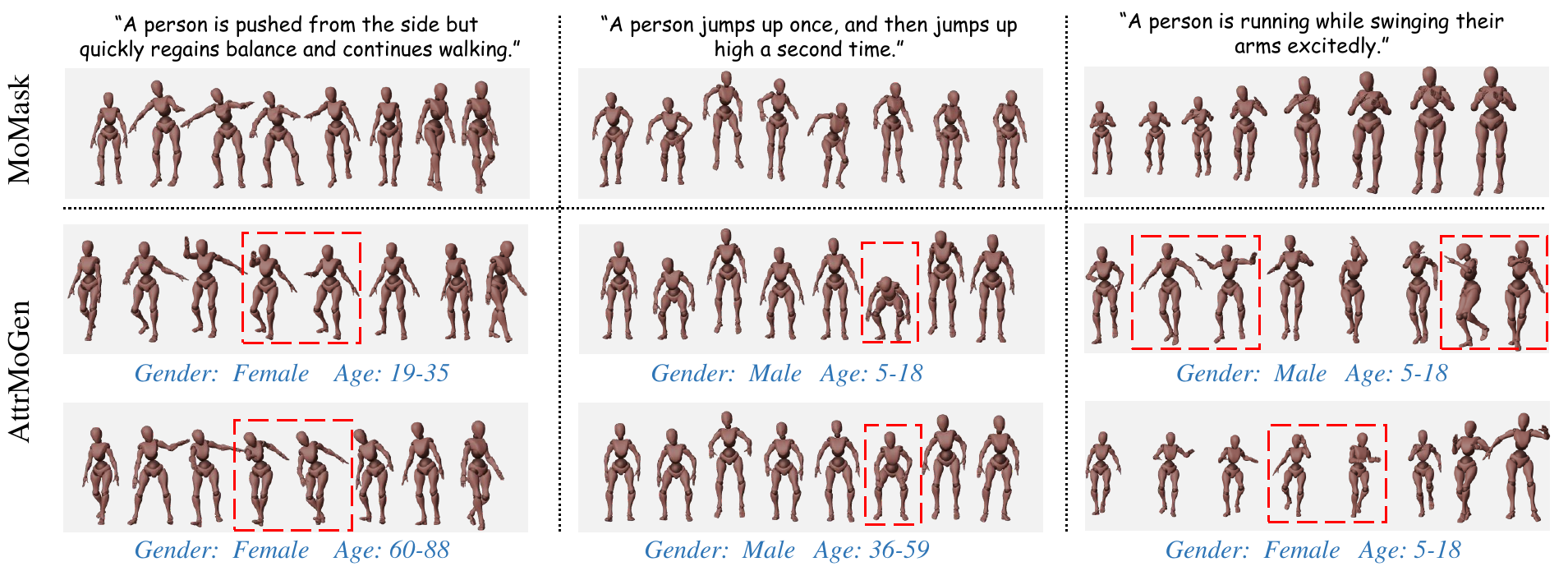}
    \caption{Visualization of generated motions of MoMask and AttrMoGen. As shown, subjects of different attributes exhibit variations in the extent and patterns of movements.}
    \label{fig:visualization}
\end{figure*}

%\label{sec:quantitative}

\paragraph{Ablation Studies.} A straightforward way to incorporate human attributes is to directly supplement attribute information into the text prompts and then employ conventional text-to-motion frameworks. Take the original MoMask model as an example, we conduct two key ablations: (1) human attributes are added to text prompts during only test time but not during training (`w/ attr test'). (2) human attributes are added to text prompts during both training and testing (`w/ attr train'). The results are illustrated in Table~\ref{tab:ablation}. As shown, directly providing attributes through text prompts during inference fails to generate high-quality motions. Including attribute information in the training texts enhances MoMask's performance, but remains less effective than our proposed attribute-control scheme.

% Modified: add hyperparameters
In addition, we analyze how each component impacts the performance of our proposed model, as well as the influence of the hyperparameters $\alpha, \lambda$ in Equation~\ref{eq:loss}. For clarity, AttrMoGen without $\mathcal{L}_{entropy}$ and $\mathcal{L}_{bottleneck}$ are denoted as `w/o entropy' and `w/o bottleneck' in the table respectively. The results are presented in Table~\ref{tab:ablation}.

\paragraph{Quantitative Evaluations.} 
% Modified: add style motion model
A number of representative text-to-motion methods are evaluated on the HumanAttr dataset, including approaches based on VAE, autoregressive models, masked transformers, and diffusion. Additionally, we adopt a style-controlled motion generation model GenMoStyle~\cite{guo2024generative} as a baseline by substituting style labels with attribute labels and using text prompts to generate content codes during inference. All experiments are repeated 20 times, and the mean values along with a 95\% confidence interval are reported. The overall performance is summarized in Table~\ref{tab:overall}. As shown, our method outperforms competing methods in R-precision, FID, MM-Dist, and Diversity. Notably, AttrMoGen advances the FID of MoMask from $0.245$ to $0.089$ and the MM-Dist from $2.602$ to $2.266$, demonstrating that incorporating human attributes can significantly enhance the quality of generated motion. Table~\ref{tab:group} further compares the performance of MoMask and AttrMoGen within each attribute group, showing that AttrMoGen consistently outperforms MoMask across all groups. % Notably, the performance varies across attribute groups, reflecting the diverse motion patterns associated with each group.

The effectiveness of attribute control is validated as follows. First, we pretrain an attribute classifier on the training set. We then use this classifier to classify the attributes of generated motions. For an ideal generation model, the predicted attributes should match the input controls. We adopt two testing protocols: (a) generating motion from (text, attribute) pairs drawn from the original test set, to evaluate performance on the original data distribution; and (b) generating motion from text with randomly assigned attributes, to assess generalization beyond the original dataset. Table~\ref{tab:acc} shows that AttrMoGen demonstrates higher classification accuracy, indicating that it successfully generates motions aligned with user-specified attributes.

\begin{table}[t]
  \centering
  \scalebox{0.73}
  {
    \begin{tabular}{ccccccc}
      \toprule[1.25pt]
      \multirow{2}{*}{Attribute} & \multirow{2}{*}{Group} & \multicolumn{2}{c}{True}& \multicolumn{2}{c}{Random}  \\
        &  & MoMask & AttrMoGen & MoMask & AttrMoGen  \\
      \midrule
     \multirow{3}{*}{Gender}  & male & 0.747$^{\pm0.002}$ &  \textbf{0.992}$^{\pm0.001}$ & 0.614$^{\pm0.004}$ &  \textbf{0.992}$^{\pm0.001}$\\
       & female &  0.546$^{\pm0.003}$ &  \textbf{0.985}$^{\pm0.001}$ & 0.383$^{\pm0.004}$ &  \textbf{0.983}$^{\pm0.001}$ \\
       & avg. & 0.657$^{\pm0.002}$ &  \textbf{0.989}$^{\pm0.001}$ & 0.515$^{\pm0.004}$ &  \textbf{0.988}$^{\pm0.001}$ \\

      \midrule \midrule
\multirow{5}{*}{Age}  & 5-18 & 0.314$^{\pm0.009}$ &  \textbf{0.422}$^{\pm0.012}$ & $0.098^{\pm0.005}$ &  \textbf{0.148}$^{\pm0.007}$\\     
       & 19-35 & 0.694$^{\pm0.002}$ &  \textbf{0.728}$^{\pm0.003}$ & 0.621$^{\pm0.003}$ &  \textbf{0.661}$^{\pm0.003}$\\
       & 36-59 & 0.409$^{\pm0.009}$ &  \textbf{0.439}$^{\pm0.009}$ &  \textbf{0.106}$^{\pm0.007}$ & 0.102$^{\pm0.005}$ \\
      & 60-88 & 0.556$^{\pm0.005}$ &  \textbf{0.787}$^{\pm0.005}$ & 0.162$^{\pm0.005}$ &  \textbf{0.387}$^{\pm0.009}$\\
      & avg. & 0.619$^{\pm0.002}$ &  \textbf{0.691}$^{\pm0.002}$ & 0.460$^{\pm0.002}$ &  \textbf{0.527}$^{\pm0.002}$\\
      \bottomrule
    \end{tabular}
      
  }
    \caption{
    Attribute classification accuracy of generated motion controlled by true and random attribute labels.}
    \label{tab:acc}
\end{table}

\begin{figure}[t]
    \centering
    \includegraphics[width=1.0\linewidth]{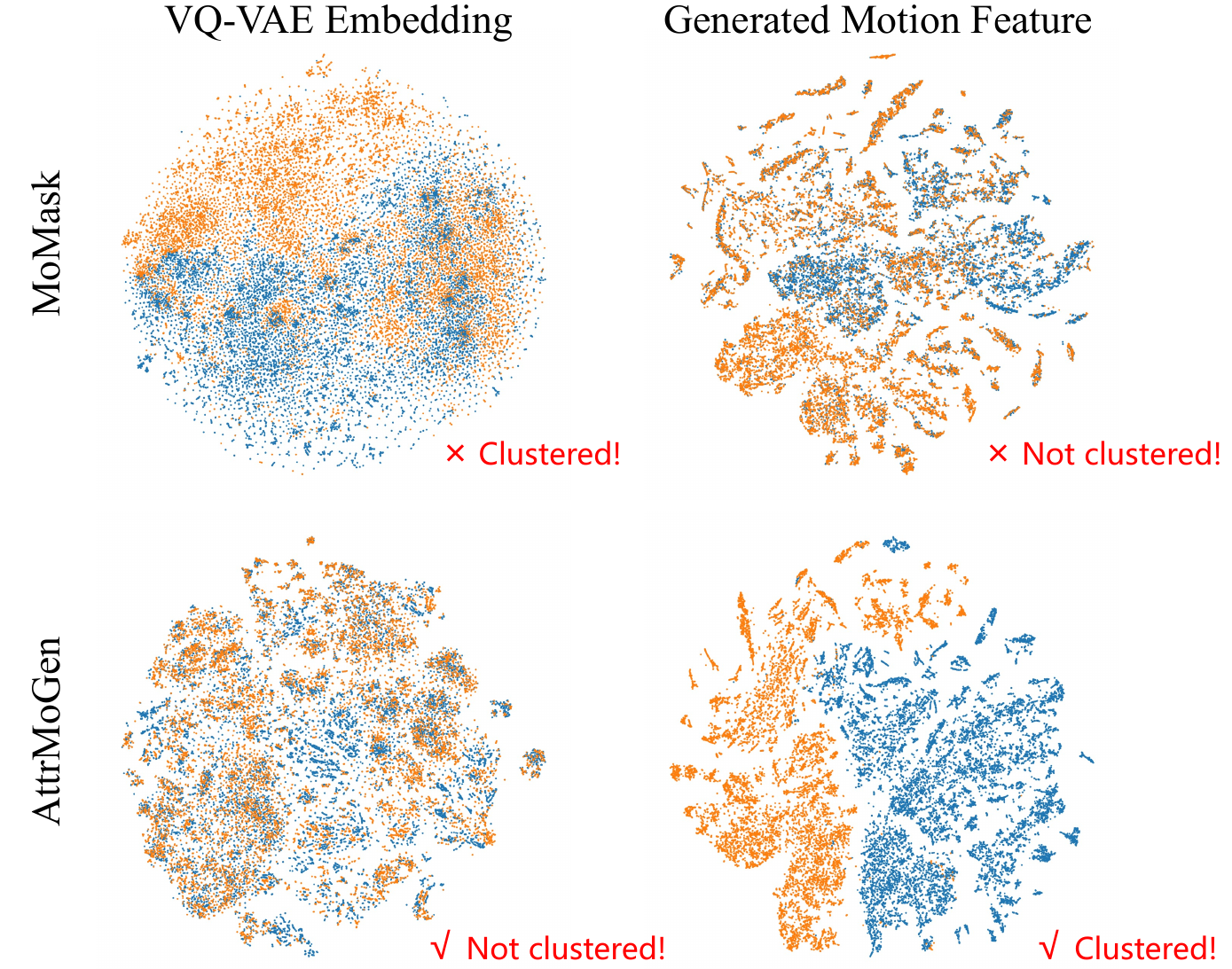}
    \caption{t-SNE visualization with colors representing male and female. The VQVAE embeddings of AttrMoGen exhibit no gender-based clustering, indicating effective removal of gender cues from the motion. Meanwhile, with attributes control input, features of the generated motion by AttrMoGen displays distinct gender-based clusters, demonstrating better alignment with attributes input. Best viewed in color.}
    \label{fig:tsne}
\end{figure}

% Modified: demo video
\paragraph{Visualizations.} To illustrate AttrMoGen's ability of attribute control, we visualized the generated motions with different attribute inputs for the same text inputs, alongside MoMask's results for comparison. As shown in Figure~\ref{fig:visualization}, the extent and patterns of movements vary across different attributes, which validate AttrMoGen's effectiveness in producing attribute-aware motions. %\textbf{Demo videos are available in the supplementary materials}.

Since the impact of gender on motion is hard to observe visually, we design a t-SNE visualization in Figure~\ref{fig:tsne}. First, motion features extracted by the VQVAE of MoMask and AttrMoGen are visualized. Unlike MoMask that showcases distinct gender-based clustering, Decoup-VQVAE removes attribute cues, resulting in no observable gender clusters. Then we feed MoMask‑ and AttrMoGen‑generated motions into a pretrained feature extractor and visualize the features. Here, AttrMoGen exhibits stronger gender‑based clustering than MoMask, indicating better alignment of generated motions with the input attributes.

\section{Conclusion}
\label{sec:conclusion}
% Modified: prune
%This work explores the novel direction of integrating human attributes into text-to-motion generation. The core design of our framework is a decoupling encoder inspired by Structural Causal Model, which separates action semantics from human attributes, enabling text-to-semantics prediction and attribute-controlled generation. We also present a text-to-motion dataset with attribute annotations called HumanAttr, which features subjects with diverse attributes. Extensive evaluations are conducted on HumanAttr, highlighting the potential of generating personalized and contextually appropriate human motions based on attribute control. Future efforts could further expand the dataset scale and the number of subjects by leveraging mocap-free data, and add more diverse types of human attributes.
This work explores the novel direction of integrating human attributes into text-to-motion generation. The core design of our framework is a decoupling encoder inspired by Structural Causal Model, which separates action semantics from human attributes, enabling text-to-semantics prediction and attribute-controlled generation. We also present a text-to-motion dataset with attribute annotations called HumanAttr, which features subjects with diverse attributes. Extensive evaluations demonstrate the potential of generating realistic human motions based on text and attribute control. %Future efforts could further expand the dataset scale and the number of subjects by leveraging mocap-free data, and add more diverse types of human attributes.

\section{Acknowledgements}
The research is supported by a grant from Bytedance (No.CT20250811106734).

\bibliography{aaai2026}

\appendix

\begin{table}[t]
  \caption{
    Results of ablation studies.
    ``$\uparrow$'' denotes that higher is better.
    ``$\downarrow$'' denotes that lower is better.}
  \centering
  \Large
  \resizebox{1.\linewidth}{!}
{
  \begin{tabular}{lccccc}
    \toprule[1.25pt]
\multirow{2}{*}{{Methods}} &\multicolumn{3}{c}{{R-Precision $\uparrow$}} & \multirow{2}{*}{{FID $\downarrow$}} & \multirow{2}{*}{{MM-Dist $\downarrow$}} \\
\cmidrule(rl){2-4}
  & Top-1 & Top-2 & Top-3 \\ 
    % TM2T~\cite{guo2022tm2t} &  0.479$^{\pm.0.002}$ & 0.686$^{\pm.0.002}$ & 0.800$^{\pm.0.002}$  & 1.491$^{\pm.0.033}$ & 3.787$^{\pm.0.014}$\\
    %  \quad w/ attr train  & $^{\pm.}$ & $^{\pm.}$ & $^{\pm.}$ & $^{\pm.}$ & $^{\pm.}$ \\
    %  \midrule
    % T2M~\cite{guo2022generating} & 0.592$^{\pm.0.002}$ & 0.778$^{\pm.0.002}$ &0.859$^{\pm.0.001}$ & 1.909$^{\pm.0.031}$ &3.827$^{\pm.0.018}$  \\
    %  \quad w/ attr train  & $^{\pm.}$ & $^{\pm.}$ & $^{\pm.}$ & $^{\pm.}$ & $^{\pm.}$ \\
    %  \midrule
    % MotionDiffuse~\cite{zhang2022motiondiffuse} & 0.670$^{\pm.0.002}$ & 0.857$^{\pm.0.002}$ &0.928$^{\pm.0.001}$ & 0.416$^{\pm.0.011}$ & 2.704$^{\pm.0.015}$ \\
    % \quad w/ attr test  & $^{\pm.}$ & $^{\pm.}$ & $^{\pm.}$ & $^{\pm.}$ & $^{\pm.}$ \\
    %  \quad w/ attr train  & $^{\pm.}$ & $^{\pm.}$ & $^{\pm.}$ & $^{\pm.}$ & $^{\pm.}$ \\
    %  \midrule
    \midrule
    MDM~\cite{mdm} & 0.330$^{\pm.0.006}$&  0.479$^{\pm.0.009}$ & 0.565$^{\pm.0.007}$&  1.581$^{\pm.0.361}$& 9.482$^{\pm.0.105}$ \\
    \quad w/ attr test  & 0.276$^{\pm.0.005}$ & 0.411$^{\pm.0.007}$ & 0.504$^{\pm.0.007}$ & 4.320$^{\pm.0.251}$ & 10.384$^{\pm.0.079}$ \\
     \quad w/ attr train  & 0.273$^{\pm.0.005}$ & 0.407$^{\pm.0.007}$ & 0.494$^{\pm.0.006}$ & 1.368$^{\pm.0.194}$ & 10.951$^{\pm.0.079}$ \\
     \midrule
    T2M-GPT~\cite{zhang2023t2m} & 0.661$^{\pm.0.002}$ & 0.824$^{\pm.0.002}$ & 0.889$^{\pm.0.002}$&  0.279$^{\pm.0.017}$& 3.203$^{\pm.0.022}$ \\
    \quad w/ attr test  & 0.498$^{\pm.0.002}$ & 0.673$^{\pm.0.003}$ & 0.758$^{\pm.0.003}$ & 1.327$^{\pm.0.039}$ & 5.374$^{\pm.0.032}$ \\
     \quad w/ attr train  & 0.655$^{\pm.0.003}$ & 0.820$^{\pm.0.002}$ & 0.885$^{\pm.0.002}$ & 0.276$^{\pm.0.016}$ & 3.274$^{\pm.0.018}$ \\
     \midrule
    MLD~\cite{mld} & 0.660$^{\pm.0.002}$ & 0.850$^{\pm.0.001}$ & 0.922$^{\pm.0.001}$ & 0.349$^{\pm.0.011}$ & 2.673$^{\pm.0.013}$ \\
    \quad w/ attr test  & 0.434$^{\pm.0.003}$ & 0.590$^{\pm.0.003}$ & 0.672$^{\pm.0.002}$ & 12.727$^{\pm.0.154}$ & 6.895$^{\pm.0.028}$ \\
     \quad w/ attr train  & 0.644$^{\pm.0.002}$ & 0.835$^{\pm.0.002}$ & 0.910$^{\pm.0.002}$ & 0.323$^{\pm.0.014}$ & 2.922$^{\pm.0.017}$ \\
    \midrule
    MoMask~\cite{guo2024momask} &0.685$^{\pm.0.003}$ & 0.864$^{\pm.0.002}$ & 0.925$^{\pm.0.001}$ & 0.245$^{\pm.0.009}$ & 2.602$^{\pm.0.009}$ \\
    \quad w/ attr test  & 0.603$^{\pm.0.002}$ & 0.781$^{\pm.0.002}$ & 0.853$^{\pm.0.002}$ & 0.957$^{\pm.0.019}$ & 3.815$^{\pm.0.014}$ \\
     \quad w/ attr train  &0.689$^{\pm.0.002}$ & 0.872$^{\pm.0.002}$ & 0.933$^{\pm.0.002}$& 0.203$^{\pm.0.007}$ &2.518$^{\pm.0.012}$ \\
    \midrule
    \textbf{AttrMoGen} &  \textbf{0.705}$^{\pm.0.002}$ &   \textbf{0.882}$^{\pm.0.002}$& \textbf{0.940}$^{\pm.0.001}$  &  \textbf{0.089}$^{\pm.0.003}$&  \textbf{2.266}$^{\pm.0.012}$  \\
  \bottomrule
  \label{tab:ablation}
  %\vspace{-1.5em}
  \end{tabular}}
\end{table}

\begin{figure}[t]
    \centering
    \includegraphics[width=\columnwidth]{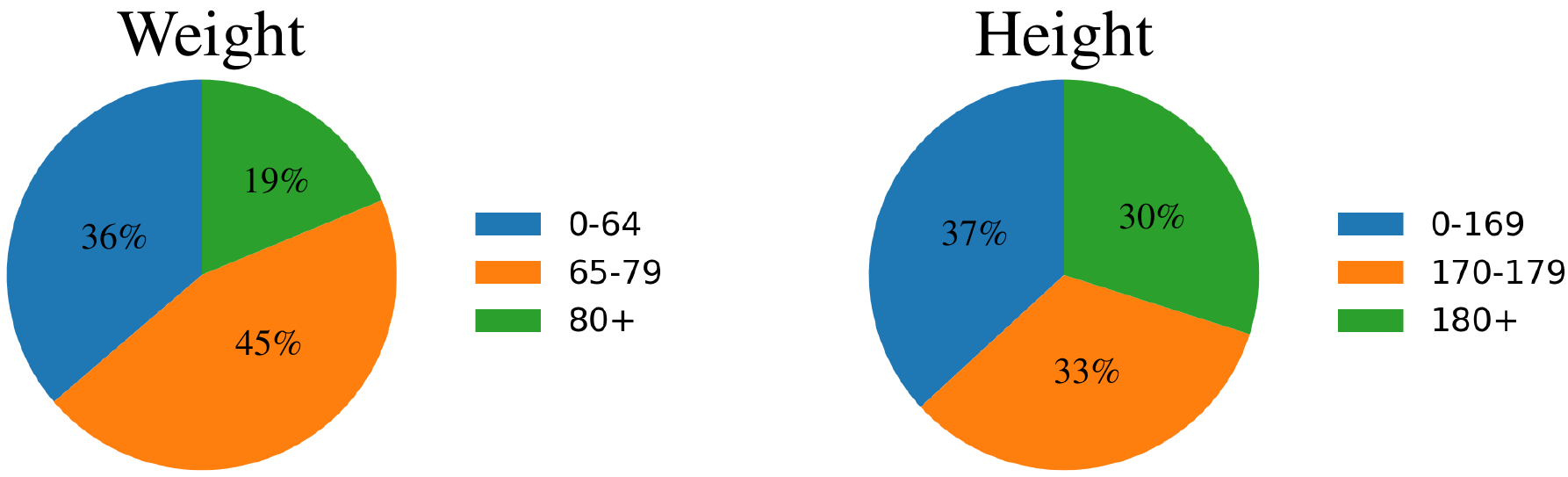}
    \caption{Statistics of weight (in kg), height (in cm) of the HumanAttr dataset.}
    \label{fig:dataset}
\end{figure}

\begin{table*}[t]
  \caption{
    Performance comparison on different attribute groups.
    ``$\uparrow$'' denotes that higher is better.
    ``$\downarrow$'' denotes that lower is better.
    ``$\rightarrow$'' denotes that results are better when closer to the real motion.}
  \centering
  \resizebox{1.0\linewidth}{!}
  {
    \begin{tabular}{cccccccccc}
      \toprule[1.25pt]
      \multirow{2}{*}{{Attribute}} &\multirow{2}{*}{{Group}} & \multirow{2}{*}{{Method}} & \multicolumn{3}{c}{{R-Precision $\uparrow$}} & \multirow{2}{*}{{FID $\downarrow$}} & \multirow{2}{*}{{MM-Dist $\downarrow$}} &Diversity $\rightarrow$ \\
      \cmidrule(rl){4-6}
       & & & Top-1 & Top-2 & Top-3 & & &  Generated / Real\\
      \midrule
      \multirow{6}{*}{Weight (kg)}  & \multirow{2}{*}{0-64} & MoMask & 0.644$^{\pm.0.005}$ &0.835$^{\pm.0.005}$ & 0.912$^{\pm.0.003}$ &0.203$^{\pm.0.011}$ &2.538$^{\pm.0.023}$ & 17.674$^{\pm.0.122}$ / 18.329$^{\pm.0.244}$  \\

       & & AttrMoGen & \textbf{0.652}$^{\pm.0.005}$ & \textbf{0.850}$^{\pm.0.004}$ & \textbf{0.920}$^{\pm.0.002}$ & \textbf{0.126}$^{\pm.0.007}$ & \textbf{2.388}$^{\pm.0.014}$ &  \textbf{18.062}$^{\pm.0.136}$ / 18.329$^{\pm.0.244}$\\
        \cmidrule(rl){2-9}
       
       & \multirow{2}{*}{65-79} & MoMask & 0.653$^{\pm.0.003}$ & 0.840$^{\pm.0.003}$& 0.914$^{\pm.0.003}$  &  0.184$^{\pm.0.013}$ &2.379$^{\pm.0.015}$  &   17.049$^{\pm.0.183}$ / 17.471$^{\pm.0.153}$ \\
       
       & & AttrMoGen & \textbf{0.668}$^{\pm.0.004}$ &  \textbf{0.854}$^{\pm.0.003}$& \textbf{0.925}$^{\pm.0.003}$  & \textbf{0.107}$^{\pm.0.007}$ & \textbf{2.249}$^{\pm.0.018}$ &  \textbf{17.165}$^{\pm.0.200}$ / 17.471$^{\pm.0.153}$ \\

        \cmidrule(rl){2-9}

       & \multirow{2}{*}{80+} & MoMask & \textbf{0.646}$^{\pm.0.007}$ &\textbf{0.837}$^{\pm.0.005}$ & \textbf{0.913}$^{\pm.0.004}$ &0.332$^{\pm.0.020}$ & 2.439$^{\pm.0.024}$ &  17.620$^{\pm.0.148}$ /  17.962$^{\pm.0.184}$\\
       
       & & AttrMoGen & \textbf{0.646}$^{\pm.0.007}$ &0.835$^{\pm.0.005}$ & 0.907$^{\pm.0.003}$ & \textbf{0.244}$^{\pm.0.011}$&\textbf{2.437}$^{\pm.0.025}$  & \textbf{18.045}$^{\pm.0.211}$ /  17.962$^{\pm.0.184}$\\
               \midrule \midrule

      \multirow{6}{*}{Height (cm)}  & \multirow{2}{*}{0-169} & MoMask & 0.633$^{\pm.0.004}$ & 0.830$^{\pm.0.005}$ & 0.906$^{\pm.0.004}$ & 0.212$^{\pm.0.012}$&2.562$^{\pm.0.018}$ &  18.181$^{\pm.0.150}$ / 18.630$^{\pm.0.242}$ \\
       
       & & AttrMoGen & \textbf{0.652}$^{\pm.0.005}$ &\textbf{0.842}$^{\pm.0.003}$ & \textbf{0.913}$^{\pm.0.003}$ &\textbf{0.112}$^{\pm.0.010}$ & \textbf{2.400}$^{\pm.0.018}$ &  \textbf{18.408}$^{\pm.0.218}$ / 18.630$^{\pm.0.242}$ \\
        \cmidrule(rl){2-9}
        
       & \multirow{2}{*}{170-179} & MoMask & 0.652$^{\pm.0.006}$ &0.843$^{\pm.0.004}$ &  0.915$^{\pm.0.003}$& 0.241$^{\pm.0.012}$& 2.406$^{\pm.0.021}$  & 16.981$^{\pm.0.131}$ /  17.269$^{\pm.0.194}$ \\
       
       & & AttrMoGen & \textbf{0.664}$^{\pm.0.004}$ &\textbf{0.855}$^{\pm.0.003}$ &  \textbf{0.923}$^{\pm.0.003}$& \textbf{0.166}$^{\pm.0.008}$&\textbf{2.330}$^{\pm.0.015}$  & \textbf{17.006}$^{\pm.0.153}$ /  17.269$^{\pm.0.194}$ \\

        \cmidrule(rl){2-9}
        
       & \multirow{2}{*}{180+} & MoMask &  0.636$^{\pm.0.006}$ & 0.822$^{\pm.0.005}$&  0.897$^{\pm.0.004}$ &0.176$^{\pm.0.013}$ & 2.395$^{\pm.0.019}$ & 16.765$^{\pm.0.257}$ /  16.924$^{\pm.0.197}$ \\
       
       & &  AttrMoGen & \textbf{0.639}$^{\pm.0.005}$ & \textbf{0.828}$^{\pm.0.004}$& \textbf{0.904}$^{\pm.0.004}$ &\textbf{0.123}$^{\pm.0.012}$ & \textbf{2.267}$^{\pm.0.023}$  & \textbf{17.040}$^{\pm.0.166}$ / 16.924$^{\pm.0.197}$\\
       
        \midrule \midrule

      \multicolumn{2}{c}{\multirow{2}{*}{Overall}}  & MoMask &  0.666$^{\pm.0.003}$ & 0.849$^{\pm.0.002}$&  0.917$^{\pm.0.001}$ &0.169$^{\pm.0.008}$ & 2.465$^{\pm.0.008}$ & 17.499$^{\pm.0.183}$ / 17.992$^{\pm.0.205}$ \\
       
        & &  AttrMoGen & \textbf{0.675}$^{\pm.0.004}$ & \textbf{0.857}$^{\pm.0.002}$& \textbf{0.924}$^{\pm.0.002}$ &\textbf{0.116}$^{\pm.0.004}$ & \textbf{2.332}$^{\pm.0.010}$  & \textbf{17.666}$^{\pm.0.175}$ /  17.992$^{\pm.0.205}$ \\
       
      \bottomrule
        \label{tab:group}
        %\vspace{-1.5em}
    \end{tabular}
  }
\end{table*}

\begin{figure}
    \centering
    \includegraphics[width=1.0\linewidth]{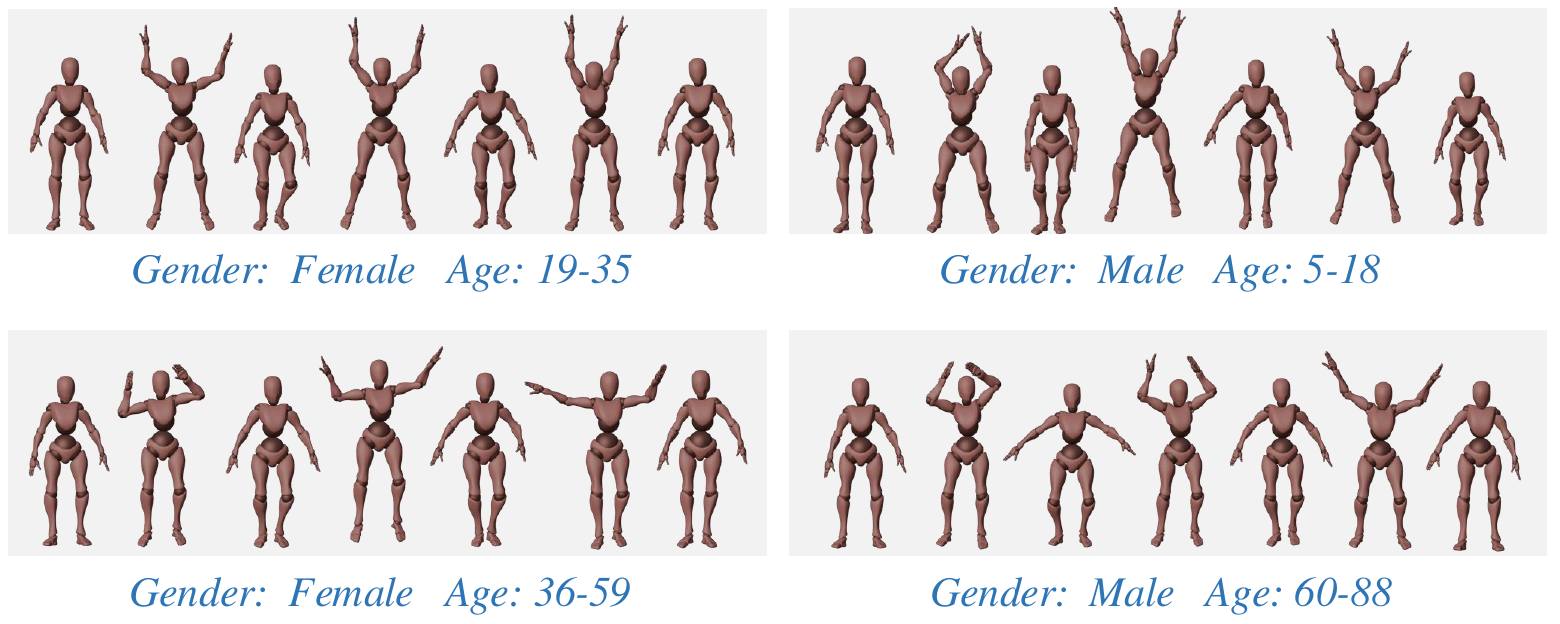}
        \vspace{-0.5em}
    \caption{Counterfactual motions visualization. The upper-left figure depicts the original motion while the remaining are counterfactuals decoded from same semantics but different attributes.}
    \label{fig:counterfactual}
    \vspace{-0.2in}
\end{figure}

\begin{figure*}[t]
    \centering
    \includegraphics[width=1.0\textwidth]{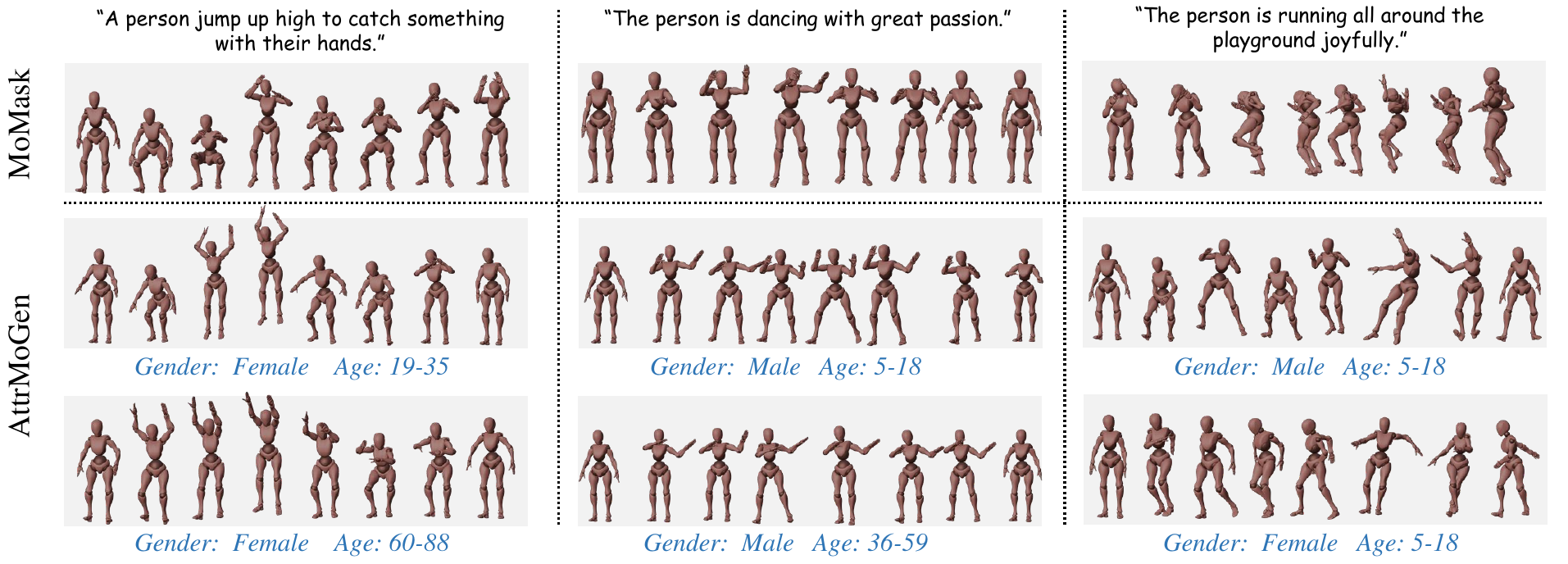}
    \caption{Visualization of generated motions of MoMask and AttrMoGen, which demonstrates AttrMoGen's effectiveness in producing realistic attribute-aware motions.}
    \label{fig:visualization}
\end{figure*}

\section{Additional Proof}
\label{sec:proof}
This section presents a mathematical proof for Equation 9 in the main text. Our goal is to derive an upper bound for the mutual information $I(X; S)$ as follows:
\begin{equation}
    I(X;S) \le \mathbb{E}_{x \sim p(X)} \mathbb{E}_{x' \sim p(X)} \mathcal{D}_{KL}(p(S|x) \|  p(S|x')).
\end{equation}
By Jensen's inequality, for any real-valued convex function $\phi$ and any probability density function $f$ satisfying $\int_x f(x)dx = 1$, the following inequality holds:
\begin{equation}
    \phi (\int_x g(x)f(x)dx) \le \int_x \phi(g(x)) f(x) dx.
\end{equation}
Note that $-\log$ is a convex function. For any given $s$, let $g(x) = p(s|x)$, $f(x) = p(x)$, and $\phi(x) = -\log x$. Applying the above inequality, we have:
\begin{equation}
    -\log (\int_{x} p(s|x)p(x)dx) \le \int_{x} -\log(p(s|x)) p(x) dx, \forall s,
\end{equation}
which is equivalent to
\begin{equation}
    \log (\mathbb{E}_{x \sim p(x)} p(s|x)) \ge \mathbb{E}_{x \sim p(x)} \log(p(s|x))), \forall s.
    \label{eq:ineq}
\end{equation}
Utilizing the above inequality, we derive:
\begin{align}
    I(X;S) &= \mathbb{E}_{x \sim p(X)} \mathcal{D}_{KL}(p(S|x) \| p(S)) \nonumber\\
           &= \mathbb{E}_{x \sim p(X)} \mathcal{D}_{KL}\big(p(S|x) \| \mathbb{E}_{x' \sim p(X)} p(S|x')\big) \nonumber\\
           &= \mathbb{E}_{x \sim p(X)} \mathbb{E}_{s \sim p(S|x)} \log \frac{p(s|x)}{\mathbb{E}_{x' \sim p(X)} p(s|x')} \nonumber\\
           &= \mathbb{E}_{x \sim p(X)} \mathbb{E}_{s \sim p(S|x)} \big(\log p(s|x) \nonumber \\
                    & \quad \quad \quad \quad \quad \quad  \quad  - \log (\mathbb{E}_{x' \sim p(X)} p(s|x'))\big) \nonumber\\
           &\le \mathbb{E}_{x \sim p(X)} \mathbb{E}_{s \sim p(S|x)} \big(\log p(s|x) \nonumber \\
                     & \quad \quad \quad \quad \quad \quad  \quad  - \mathbb{E}_{x' \sim p(X)} \log p(s|x')\big) \nonumber\\
           &= \mathbb{E}_{x \sim p(X)} \mathbb{E}_{s \sim p(S|x)} \mathbb{E}_{x' \sim p(X)} \log \frac{p(s|x)}{p(s|x')} \nonumber\\
           &= \mathbb{E}_{x \sim p(X)} \mathbb{E}_{x' \sim p(X)} \mathbb{E}_{s \sim p(S|x)} \log \frac{p(s|x)}{p(s|x')}  \nonumber\\
           &= \mathbb{E}_{x \sim p(X)} \mathbb{E}_{x' \sim p(X)} \mathcal{D}_{KL}(p(S|x) \| p(S|x')).
\end{align}
Thus we reach the desired result.

\section{Additional Details}
The dataset is split into training, validation and testing sets in an 80\%, 5\%, and 15\% ratio. All data preprocessing protocols follow those used in HumanML3D~\cite{humanml3d}, and the input dimension for the encoder is 263. For the model architecture, MoMask~\cite{guo2024momask} is adopted due to its superior performance and efficiency. This architecture consists of a residual VQVAE, a masked transformer, and a residual transformer. The network architecture for the proxy classifier $h$ is the same as VQVAE encoder. To ensure a fair comparison, all experimental settings and hyperparameters (including seed, optimizer, epochs, scheduler, batch size, codebook size, etc.) strictly follow those in MoMask~\cite{guo2024momask}. Training is performed on a single RTX 4090 GPU, requiring 8 hours for the VQVAE and 26 hours for the generative transformer.

For the evaluation metrics, we follow the standard protocol~\cite{humanml3d} and report the following metrics: 
(1) Frechet Inception Distance (FID)~\cite{heusel2017gans}, measuring the distance between feature distributions of generated and real motions; (2) R-Precision, the accuracy for motion-to-text retrieval in a batch of 32 samples; (3) Multimodal Distance (MM-Dist), measuring the Euclidean distance between the feature vectors of text and motion; (4) Diversity, which calculates the average Euclidean distance between features of 300 randomly sampled motion pairs; (5) Multimodality, assessing the diversity of generated motions for the given text. The feature extractor used for evaluation is pre-trained on the HumanAttr dataset, following the same method as in~\cite{humanml3d}.

Regarding how action labels in certain sub-datasets are expanded into full text, there are two cases: (a) detailed descriptions of movements for each action class are provided in the database, which can be directly used (K3DA); (b) action labels themselves are already descriptive phrases; we simply add appropriate subjects to form complete sentences (ETRI, Kinder-gator).

\section{Additional Experimental Results}
To investigate the effect of directly incorporating attribute information into text prompts, we conduct experiments across several baselines~\cite{mdm, zhang2023t2m, mld, guo2024momask}.
In Table~\ref{tab:ablation}, `w/ attr test' indicates that human attributes are added to text prompts during only test time but not during training, while `w/ attr train' indicates that human attributes are added to text prompts during both training and testing. The results demonstrate that including attribute information during training generally improves the FID scores for most baselines. However, this approach remains less effective compared to our proposed AttrMoGen.

Also, an additional experiment is conducted on a data subset containing human weight and height annotations, following the standard evaluation protocol. For attribute control, weight (in kg) is categorized into three groups: 0-64 as group 0, 65-79 as group 1, and 80+ as group 2. Height (in cm) is similarly divided into three groups: 0-169 as group 0, 170-179 as group 1, and 180+ as group 2. Figure~\ref{fig:dataset} illustrates the distribution of weight and height within the HumanAttr dataset. 
The results, presented in Table~\ref{tab:group}, compare the overall performance of MoMask and AttrMoGen, as well as their performance within each attribute group, showing that AttrMoGen consistently outperforms MoMask.

\section{More Visualizations}
We provide more visualizations of the generated motions with different attribute inputs for the same text prompts, as well as the results from MoMask. As illustrated in Figure~\ref{fig:visualization}, AttrMoGen effectively generates realistic, attribute-aware motions across varying human attributes.
% Modified: counterfactual motions vis
Figure~\ref{fig:counterfactual} further demonstrates the quality of counterfactual motions in the Decoupling VQVAE's information bottleneck.

\end{document}